# Accessible Survey of Evolutionary Robotics and Potential Future Research Directions

Hari Mohan Pandey
Data Science and Artificial Intelligence Department, Bournemouth University, United Kingdom
profharimohanpandey@gmail.com

**Abstract:** This paper reviews various Evolutionary Approaches applied to the domain of Evolutionary Robotics with the intention of resolving difficult problems in the areas of robotic design and control. Evolutionary Robotics is a fast-growing field that has attracted substantial research attention in recent years. The paper thus collates recent findings along with some anticipated applications. The reviewed literature is organized systematically to give a categorical overview of recent developments and is presented in tabulated form for quick reference. We discuss the outstanding potentialities and challenges that exist in robotics from an ER perspective, with the belief that these will be have the capacity to be addressed in the near future via the application of evolutionary approaches. The primary objective of this study is to explore the applicability of Evolutionary Approachess in robotic application development. We believe that this study will enable the researchers to utilize Evolutionary Approaches to solve complex outstanding problems in robotics.

1. Introduction

In the last few years, bio-inspired computational techniques [1] [2] [3] [4] have demonstrated considerable impact upon robotics and therefore, the interest of the roboticists bends towards the utility of these techniques as it address the key challenges involved in complicated design and control of robots Evolutionary techniques are population based and have gained popularity in the robotics community as these techniques have ability to generate super-individual (iteratively learns the basic features from a given scenario, adapt the environment and then evolve out certain features, which are not actually present in the initial population). Sims [5] showed the feasibility to utilize an evolutionary process in designing robot morphology and robot brain respectively. The efficiency of the techniques of handling challenging problems of robotics has opened up the field of Evolutionary Robotics (ER) and a deeper understanding of these techniques would enable further development. Algorithms in ER are usually work upon a population set (initially selected from a random distribution). The population set is then repeatedly updated according to the fitness function, which is formulated to express as the objective function of a problem. Evolutionary Algorithms (EAs) offer various features that are required for different aspects of non-continuity and multi-modality and hence EAs serve as a strong platform to work upon for robotics application. The most commonly used EAs in the field of the ER are Genetic Algorithms (GAs) and Genetic Programming (GP).

ER is utilized to develop a suitable control system of the robot using artificial evolution. Previous scientific literature reveals that evolution (capture slow environmental change occur after several generations) and learning (adaptive changes in an individual during its lifetime) are the two different form of biological adaption that follows different timeline. The internation between evolution and learning of artificial evolution (for example GA for evolution and Neural Network (NN) for learning) are studied extensively for several applications. ER successfully deals with this interation.

In behaviour-based robotics, a task is categorized into a number of basic behaviours. Each basic behaviour is implemented in layered fashion (separate layer of the robot control system) [1]. In this scenario, the control system is developed incrementally layer-by-layer, where each layer is responsible for a single basic behaviour. Oftentimes, trial and error approach is utilized as a coordination mechanism of the basic behaviours and coordinated through a central mechanism. Here, we noticed that the number of layers are directly proportional to the complexity of the problem. Sometimes, the number of layers go beyond the capability of the designer. Hence, a designer faces difficulty to define all the layers, their interrelationships and dependencies. This discussion concludes that there is a huge demand of an appropriate technique by which the robot coulde acquire behaviours automatically according to the situations of changining environment. ER provides a feasible solution to this problem. ER avoids direct involvement of the designer (plays a passive role in this scenario) and the basic behaviours emerge automatically applying evolution based on the interactions between the robot and its environment. Hence, decomposion of a task into several basic behaviours with their coordination are performed by a self-orgnaizing process. A fitness function (or objective function) is defined that measures the ability of a robot performing a desired task. Nelson et al. [94] has discussed various fitness function that can be used in ER. The robot and the environment creates a highly dynamical system. In this situation robot's decision at a specific time depends on the sensory information and its previous actions [96].

In many case, the behaviour of the robot controller may not be optimal initially, but after incorporating ER (for example GA) the robot controller improves its performance and yields better results. Hence, a robot controller evolves by a self-organizing process. In ER, the environment is important and it plays a central role. The environment determines a suitable basic behaviour which is active at any given time. The behaviour of a robot is influenced by its interaction with environment. A simple robot shows much complex behaviour due to dynamic internation of robot control system with the environment. In such situation, it is hard to predict the type of behaviour that are produced through the interaction between robot control system and environment. In addition, prediction of robot's control system that might yield a desired behaviour is also difficult. This discussion reveals that design of the control system is a challenging and it can be addressed through adaptive behaviour (achieved by evolution). Previous scientific research indicates the utility of the ER to address these challenges in a more effective manner as compared to behaviour-based robotics. ER avoids explicit design mechanism and develop an effective coordination through a self-orgnaizing process. ER utilizes techniques, such as GA [12], GP [97] and evolutionary strategy (ES) [98] to evolve the robot controller design, robot design, morphology and control, physical prototyping of robot. These techniques are based on population of different genotypes (information that evolves over successive generations), which is used to code robot's architecture and behaviour (phenotype), is created at random. Robots are allowed to interact with the environment and fitness function value is assigned to each of them. Robots with higher fitness function value are allowed to reproduce (through crossover and mutation) better solutions through iterations.

ER is one of the popular field of robotics research, which seems to have enough potential to develop an intelligent and autonoumous robot. It is a challengining area with several open research problems. The primary interest of this paper is to present a collection of works proposed in the area of the ER, which are then classified for the quick representation and review and finally we present a discussion and possible directions for future research.

This paper is organized as follows: Section 2 discusses the various computational tools that are utilized largely for robots. Background and various achievements are presented in Section 3. The summary of reviewed literatures is presented in Section 4. Section 5 shows the discussion and future direction for research followed by concluding remarks in Section 6.

## 2. Overview of computational tools

In this section, we present a brief summary of the popular techniques in adaptive robotics such as GA, Neural Network (NN), NeuroEvolution of Augmenting Topologies (NEAT) and Generative Representation in ER.

*2.1 Genetic algorithms*

GA is a search and optimization algorithm belongining to the class of EAs. It has demonstrated the ability to solve optimization problems using the techniques of natural evolution [12] [13] [14] [15] [16] [17] [18]. It works on population of solutions each of which is assigned a fitness value. GA use crossover, mutation operators, which can update the population. GA searching is iteratively guided by the fitness function value of the current parent generation. Whenever, we apply a GA to an optimization problem, it is typical to apply a large set of individuals (typically in the thousands), where each individual represents a solution. The solutions are recombined to create an offspring. Exisiting scientific literatures [12] [15] [74] [75] have demonstrated that information of previous generations are only implicitly and partially preserved in the current generation. Therefore, it is difficult to manage due to several reasons [76]. A key challenge with a search and optimization technique is to avoid premature convergence – a situation when diversity of the population decreases over generations [76]. A GA suffers with premature convergence. Recombination operators (crossover and mutation) and selection process are utilized to alleviate premature convergence. Several algorithms [77] [78] [79] have been proposed to avoid premature convergence in GAs. GAs are very popopular for robotic applications. It is utilized for various controller design, path finding, obstacle avoidance etc.

*2.2 Neural Networks*

NNs are an information processing model. It is inspired by the biological nervous system and it processes information similar to the brain [12] [13] [14] [15] [16]. The performance of NNs are largely depends on the structure of information processing system, which is composed of a large number of highly interconnected processing elements referred as neurons and connected through synapse. A typical NN is a network of neurons that is similar to a human being and it learns by experience [19]. A NN is largely utilized to configure parameters, through a learning process, to achieve a specific objective such as robot's control.

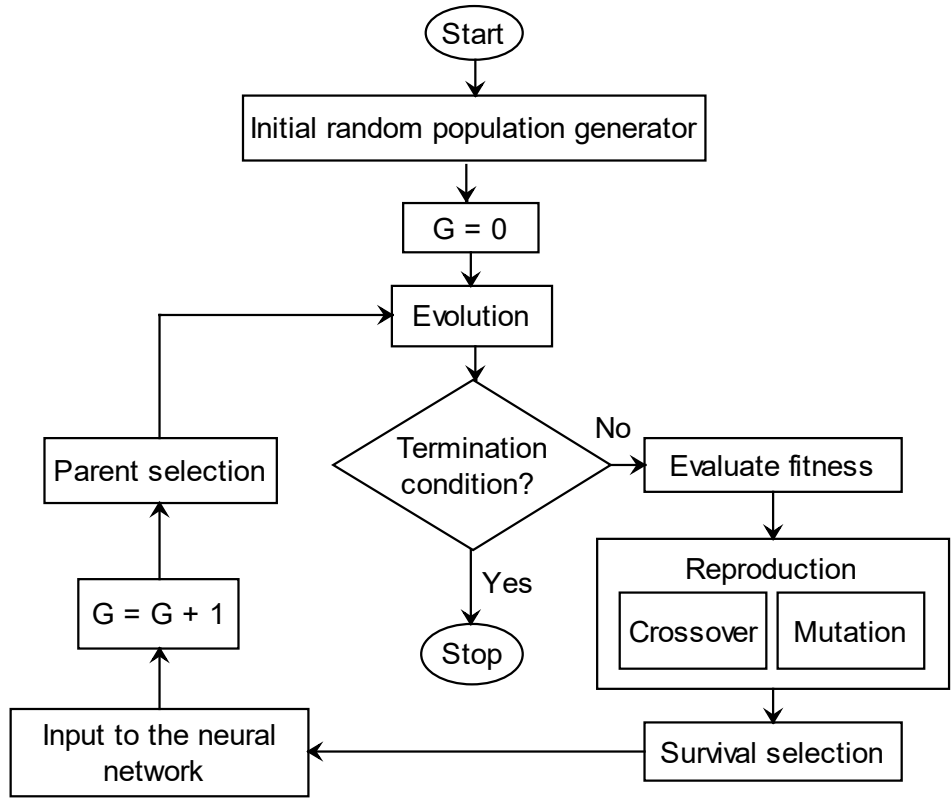

**Fig. 1**. A flowchart of GA represents the process of training a neural network.

A typical NN consists of three important layers: input layer, hidden layer and output layer. The input layer consists of sensors. It fetches the feedback signals and sends it to the hidden layer. Hidden layers are responsible to process the input signal and send the output signal to the output layer. Now, based on the output, the actuators perform actions. Figure 1 shows the training process of a NN through a GA. The training and tuning of a NN is an important step and it must be achievedfor a great performance.

*2.3NeuroEvolution of augmenting topologies*

Stanley and Miikkulainen [20] have proposed NeuroEvolution of Augmenting Topologies (NEAT). It outperformed the best fixed topology method on the challenging benchmark reinforcement learning. NEAT utilizes a GA for evolving Artificial Neural Networks (ANNs) and maintains a good balance between the fitness of evolved solution and their diversity by altering both the weighing parameters and the structure of NNs. NEAT follows three key steps:
- Tracking genes with historical markers and allows crossover operation among topologies.
- Apply speciation – the evolution of species which helps to preserve innovations.
- Developing topologies incrementally from initial structures.

NEAT offers a solution to a problem by competing conventions in a population of diverse topologies. Historical information managed by the NEAT is useful in comparing and clustering similar networks. NEAT is advantageous because it provides a cost effective analysis of mating networks. The working of NEAT is simple: it starts with an initial random population of simple NNs and then adds complexity (adding new nodes and connection via a mutation operation) over generations [21]. NEAT does not require prior information about the topology of the network. It searches through increasingly complex networks in order to get a suitable level of complexity and tend to get a solution network, which is close to the minimal required size. This step of NEAT is effective because starting with a population of minimal topologies is advantageous [21].

*2.4 Generative representations*

Generative Representation (GR) is an encoded data structure. It is a powerful representation technique can be reused recursively in creating a design. GR allows robot morphology to generate a context-free development rule from a basic "seed". This type of GR scheme was first proposed by Lindenmayer [22]. Hence, it is known as L-system. A GR scheme is an L-system program as it produces (after compilation) a sequence of build commands referred as assembly procedure. After generating assembly procedure, a GR system creates a constructor and then executes the assembly procedure, which generates both robot morphology and robot controller. A state of the art work on the GR and its application for automated design of modular robots has been presented in [23] [24] [25].

## 3. Existing Research work

Floreano and Mondada [8] utilized a GA to develop a neural controllers for autonomous robots. The evolution of controllers took place on a physical robot – *Khepera* (a miniature test mobile robot) [99]. The aim was to develop a robot controller which can detect a collision free straight path with high velocity during navigation [8]. There were eight inputs (directly taken from the sensors) and two outputs (conneted to the motors) of the NN. It was noticed that GA-based learning of the neural controller for autonomous robot was successful during navigation. Later on, Floreano and Mondada [100] utilized a GA to evolve a NN controller for a Khepera to locate a battery charger. In [100], twelve inputs (taken sensory information) and two outputs (one for each motor) of the NN controller was considered.

Miglino et al. [101] utilized a GA to evolve the weights for a NN controller. A computer simulation was conducted to train the NN controller. The best NN controller evolved through the GA was downloaded on to a real robot which produced a reasonably good results. Previous scientific research revealed that a GA was successfully utilized to evolve a feedforward neuro-controller to clean an arena surrounded by walls [102] [103] [104]. *Khepera* with a gripper module had to clean an arena, which was full of small cylindrical trash objects placed at random. The fitness function value was determined by counting the number of objects that were placed outside the arena during a given evaluation time. Miglino et al. [95] and Lund and Miglino [105] used a GA to evolve control system of a real robot. Lund and Miglino [105] mainly evolved a two layer feed forward NN with hidden units. Authors [105] accepted that evolving the whole population for a real robots for many generation is time consuming. In order to address this challenge, a simulation based approach was adapted where the main part of evolution is conducted in a simulator that reasonably reduces the time consumption. Lund and Miglino [105] described the applicability of Khepera miniature mobile robot in developing simulator with a semi-autonomous proess which reduces the gap in performance during transferring a robot control system from the simulator to the real environment.

Soloman [106] suggested a two-stage evolutionary strategies (ES) with self adaption of the step size to evolve a control architecture for Braintenberg vehicles. This ES based development of controller reasonably shown better results. In addition, this result was important, since the whole development was carried out in a real system. Soloman [106] showed that ES based controller development was found effective to speedup the development process. Theoratical reason (epistatic interaction [107]) was presented, which shows the superiority of ES over GA.

Smith [27] utilized a GA to tune the weights in a fixed NN architecture to perform a hard task, playing football match. A khepera with sixeent input and output respectively for sensory information and motor control information of the NN, which was successfully evolved for football playing. During simulation, *minimal simulation* technique was employed to ignore certain features through making the features unreliable, which helped to construct a simulated environment for a robot utilizing vision based system. In order to evaluate the performance two different scenarios were presented [27]. Firstly, robot was instructed to find a white strip in a dark arena. Secondly, robot's task was to find a tennis ball and pushing it into a goal. For both the scenario, the robot was successfully evolved same behaviours (locating the ball and scoring the goals) in simulation and in the real experiment.

Nolfi et al. [108] [109] proposed a simulation model of *phenotypic plasticity* (phenotype plasticity is used to describe the role of evolution and environmental influence on the individual [110]), where organism was modeled through a NN which was evolved through a GA. The proposed NN controller was first evolved on simulation later on implemented for a real robot

– Khepera, where a mismatch in the result (simulated and real sensory-motor implementation) was noticed.

Brooks [1] presented a robust layered control system architecture for controlling mobile robots. Koza [111] developed a layered subsumption architecture presented in [1] for simulated robots. These robots were utilized for wall-following and box-moving tasks. Reynolds [112] suggested that the behavior of a simple computational model can emerge through evolution under selection pressure from an effective fitness measure. A GP paradigm was utilized to model the evolution. An evolved program was created to determine the structure of visual sensory arrray and the mapping from sensory data to motor. In [112], fitness was measured considering the number of simulation steps the agents can run before collisions. Nordin and Banzhaf [113] also implemented GP paradigm to evolve robotic control of a real robot, which helps in obstacle avoidance and object following behaviours in Khepera. Authors [113] have applied GP to manipulate machine code that evolve control programs for robots. Through this process, the objective was to find a control program that reacts to the sensory information and produce the controlling actions. An on-line control method was proposed which was evaluated in two different environments and utilized for two tasks: obstacle avoidance and object following. The outcome of the experiments was interesting where the robot controller showed a promising results for both obstacle avoidance and object following. Lee et al. [114][115] presented an GP based approach to program behavior based robots. It was explained that evolving all the behavior controller for over all tasks is time consuming. To address this issue, Lee et al. [114] [115] presented an intermediate level based evolution through GP. It includes evolving behavior primitives and behavior arbitrators for a mobile robot – Khepera. The performance of the proposed approach was evaluated for a moderate complicated box-pushing task. Intially, the proposed approach was implemented in a simulation and later on transferred to a real robot (Khepera) without loss of performance.

The aforementioned discussion revealed that GA is a popular ER method. But, it is limited to a finite search space with fixed-length genotypes. Hence, for artificial life applications it is useful to extend GA to open-ended evolution with variable length genotypes. In this direction, researchers at the University of Sussex, UK, suggested that an improved ER is required which can show an adaptive improvement to the environmental changes rather than an optimizer. Hence, they switched over from the traditional GA to a Species Adaptation Genetic Algorithm (SAGA). The concept of SAGA was proposed by Harvey [116]. In a standard GA [12] [13], crossover operators is considered as a central searching operators whilst mutation probability is kept to a low value. The standard GA stops searching when it converges to global solution.

On the other hand, in SAGA most of the evolutionary progress is made after the convergence of the algorithm. The idea of SAGA is based on epistasis and fitness landscape which suggest that "*in the longer run a population must have genotypes of nearly equal length, and this length can only increase slowly*". In this circumstance the population will be coverge and evolves as a species. The aforementioned discussion revealed that evolution is not an optimization process rather an ongoing adaptation to the changing environments. Harvey et al. [117] [118] evolved behaviours for a gantry robot through SAGA along with Dynamical Recurrent Neural Network (DRNN). Husbands et al. [119] discussed and attempted to justify the role of aritificial evolution and ER to the development of autonomous agents with sensorimotor capabilities. Authors [119] presented a comprehensive discussion on the SAGA that shows its utility in robotics. A case study on gantry robot was shown. A concurrent evolution of visual morphologies and control network was implemented that genetically specify region of the robot's visual field to be subsampled. This process is advantageous as it provide only visual inputs to the control network. Husbands et al. [119] who argued that "*evolutionary simulations is a science*", but some gap exists (gaps are listed in [119]) in our scientific understanding of sensorimotor evolution. It was claimed that GA offers several features that can be utilized for simulating the evolutionary origins and effects of sensorimotor systems [119]. Jackobi [120] [121] developed a theoretical and no-traditional framework to identify the circumstances under which the agent performs a behaviour within an environment. Later on, the framework was utilized to prove the relaiability of an agent performing a certain behaviour within simulation ad real world, providing that certain conditions are fulfilled by both the controller and the simulation. These minimal set of conditions for behavioural transfer were practically implemented to the building of minimal simulation for ER and later on downloaded into reality [120] [121].

Colombetti and Dorigo [122] showed a parallel implementation of a learning classifier system for robot control. ALECSYS software [123] was used to evolve the control architecture of the robot (AutonoMouse - a mouse shaped autonomous robot). Authors [122] argued that "*reinforcement learning is suitable for shaping a robot*". In reinforcement learning paradigm the trainer need not present examples of correct behaviour (robots rely on its own exploration activity). But, reinforcement learning is based on feedback information that guides the learning is strongly local. Hence, many behaviours are difficult to describe in terms of reward and punishments. Similar problem was appeared in [122] when it come to measuring the robot's performance. Hence, authors [122] recommended to set up the conceptual apparatus and the technical terminology to describe the qualitative aspects of emerging behaviour of artificial agents. Grefenstette and cob [124] designed SAMUEL classifier machine learning system to explore alternative behviour in a simulated environment. SAMUEL showed ability to construct high performance rules from experience (in reinforcement learning). Later on, Grefenstette & Schultz [9] utilized SAMUEL GA learning system to evolve collision-free navigation task for mobile robots (Nomad 200). This approach was found efficient when rule sets obtained through simulations were transferred to the real robot.

Meeden [10] explored an incremental process that provide guidance to adapting controller. A local and global method of reinforcement learning was contrasted to develop a recurrent neural network (RNN) controller using GA for a simple robot (a 4-wheel robot). Author [10] showed comparative analysis of local method (complementary reinforcement back-propagation (CRBP)) and global method (GA) and suggested that these methods complement each other. A suggestion (use hybrid (GA + CRBP)) was made "*use GA to find a good starting point in the weight space and then use CRBP to do the fine-tuning*". The robot developed in [10] had to avoid contact with the walls during the movement and either to seek or avoid leight depending on the goal. The conditions were considered to determing the fitness function value of the GA. The neural network utilized during implementation consists of seven input, five hidden units and four outputs.

Ram et al. [4] showed the applicability of GA to the learning of local robot navigation behaviour for reactive control systems. The navigation problem was of a robot was divided into several basic behaviours such as move-to-goal, avoid-static-obstacle etc.

Ram *et al* (1994) proposed one method in which navigation problem of a robot was divided into some basic behaviours, namely move-to-goal, avoid-static-obstacle, and others. These behaviours were implemented with the help of some parameters, such as goal gain (strength with which a robot moves towards its goal), obstacle gain (strength with which a robot moves away from obstacles) and obstacle sphere of influence (distance from obstacle at which a robot is repelled). They used a GA to find the suitable combination of these parameters so that the robot could find a collision-free path during its navigation.

Baluja (1996) developed an evolutionary algorithm named population-based incremental learning (PBIL) for designing a neural controller. The performance of the controller was tested on Carnegie Mellon's *NAVLAB autonomous land vehicle* for its steering control and found to be satisfactory.

Pratihar *et al* (1999a) developed optimal fuzzy logic controller by using a GA-based tuning off-line. The GA-tuned fuzzy logic controller was efficient in planning optimal collision-free path of the robot, while navigating in the presence of some moving obstacles. The algorithm was tested on simulations but is yet to be tried on experiments with real robots.

Jeong & Lee (1997) developed a two-stage controller for two-wheeled soccer-playing robots using a GA. In the first stage, some rules were evolved whose condition parts involve the positions of the ball, opponents, partners and goal, whereas the action parts indicate the actions to be taken, such as a move, a dribble or a kick. In the second stage, optimal on-off type control signals were produced which allowed a robot to reach a position with desired coordinates and orientation.

Beer & Gallagher (1992) used a standard GA to determine time constants, thresholds and connection weights of a continuous-time fully-connected recurrent neural network. The locomotion control of a six-legged insect-like robot was studied in their work. The effectiveness of their method was tested on both simulation and experiment with real robots.

Galt *et al* (1997) successfully derived the optimal gait parameters of an 8-legged walking and climbing robot using a GA. The gait parameters, namely phase (time relationships between the legs) and duty factor (support period of a leg) were encoded in the GA. The evolved controllers were found to perform well in generating suitable gaits of the 8-legged robot.

Gruau & Quatramaran (1997) evolved neural controllers using cellular encoding. They were successful in generating a quadruped-locomotion gait of an *OCT-1* robot. Gomi & Ide (1997) used a set of 50 control algorithms (software) to evolve a suitable gait of an 8-legged *OCT-1* robot. Each of these control algorithms was tested for generating gaits during a fixed amount of time. The robot is required to stand up and generate its gait to move forward. The program was run for a few generations and a mixture of tetrapod and wave gaits was obtained.

More recently, Pratihar *et al* (1999b, 2000) developed optimal/near-optimal fuzzy logic controllers (FLCs) for generating suitable gaits of a six-legged robot using a GA. The hexapod is supposed to cross a ditch or take a circular turn, keeping the minimum number of legs on the ground with maximum average kinematic margin. Each leg of the six-legged robot is controlled by a separate FLC. The GA-based tuning of the FLCs is done off-line. Thus, this algorithm is suitable for on-line implementations. The effectiveness of the algorithm is tested through computer simulations and found to be satisfactory.

## 4. Classification of work

Since the seminal work conducted by Sims [5] the field of ER has evolved in three main directions, namely, Evolutionary Robotics Control (ERC), Evolutionary Robotics Design (ERD) and Co-evolution of Morphology and Brain (CEMB). Along with these, there is one more field known as Evolution of Robot Prototype (ERP), which has emerged in the past few years. This section focuses on the utility of the ER in different areas.

*3.1 Evolutionary robotics in controller design*

Artificial evolution has played an important role to evolve the robot brain. In general, NNs are utilized to teach a robot about various tasks in different exemplary situations and then the robot is expected to take decisions in an unknown situation depending upon the knowledge induced. In case of NNs, weights are used to determine the strength of a signal in a layered manner (one layer to the next one). These weights are adjusted and tuned according to the requirements of the controller. EAs, such as GAs are implemented widely to the NN controller that determines the tunned weights for the controller. EAs was applied to evolve NNs for adaptive controller parameters such as time constant, thresholds and connection weights for a 6-legged insect like robot [2]. Ram et al. [4] implemented a GA to produce a collision free path for a wheeled robot. In [4], the navigational problem was classified into basic behaviors: move-to-goal, avoid-static-obstacles, stay-on-path and others. The GA was utilized to implement these behaviors to evolve collision free paths. Authors [9] had also addressed the collision avoidance navigation problem for a Nomad 200 mobile robot in the presence of moving obstacles.

During 1990, the ER methods started gaining popularity in automatic evolution of the controller. Meeden [10] proposed a controller for a 4-wheeled robot. It avoids contact from walls through a light source as when needed utilizing fitness function. On the other hand, Baluja [28] proposed an EA based neural controller using a population based incremental learning (PBIL), which is tested on steering control of a mobile robot.

Jeong and Lee [29] implemented a 2-stage controller for 2-wheeled soccer playing robots by utilizing a GA. The first stage was responsible for the position of the ball, the opponents, the partners and the goal along with various actions (move, dribble or kick), whilst optimizing on off signals are considered in the second stage. There exist literatures present have shown the utility of the ERs – are implemented to acquire different features of the controller for Khepera (a mature mobile robot) [8] [11] [26] [27]. There are several other work in the scientific literature exist that presents the applicability of the ER for the controller designs of various robotic mechanisms [30] [32] [33] [34] [35] [40] [41] [42] [63]. Table I presents the summary of achievements in the robotic controller design through different ER techniques. It is evident that ERs address various challenging tasks of robot control. However, although, research work conducted in the area of ER has achieved many key milestones, it is clear the field is still open for further improvement, related to complex mechanisms, dynamic environment etc.

*3.2 Evolutionary robotics in design*

Existing research revealed that selecting a specific morphology for a given set of tasks is challenging and consequently a key area of research to focus on. Many researchers have utilized evolutionary computation (EC) techniques to address the robot morphology design problem. A task-based design of modular manipulator was proposed [44]. It uses 2-level GA, where layer-1 (upper layer) evolves the robotic topology whilst layer-2 (lower layer) is responsible to search for a subtask configuration. Chung et al. [45] used a GA and suggested an optimal determination of the link lengths of modular manipulator. Funer and Pollack [46] [47] suggested different 2D/3D morphologies using EC techniques. Morphologies were selected depending upon artificial selection algorithms through LEGO components. Authors [25] [48] [49] proposed grammar-like rules for generating physical systems, which was further utilized for 2D locomotive robots. The merits of a generative design specification are: it can reuse components by providing the ability to create complex modules. Table II presents various significant works that demonstrate the applicability of the ER in robotic design. More such applications might be developed in the near future using ER techniques for complicated problems. We believe that a much deeper understanding about ER is required, which can be applied to design an autonoumous manipulators for a given environment.

*3.3 Evolutionary robotics in co-evolution of morphology and brain*

Co-evolution morphology and brain in robot was introduced of Sims in 1994 [5] [62] and since then it has gained significant attention of several researchers. Sims [5] [62] demonstrated that a number of aspects related to one physical life form can be utilized to obtain an optimized life-form. In this direction, a hybrid approach using GA and GP was proposed to evolve both the controller and the robot bodies simultaneously to perform behavior specified tasks. Table III presents the research conducted in the direction of co-evolution of robot morphology and control. It is evident from Table III that main thrust of research work is to investigate the usability of the ER for gait development and motion planning for some specific locomotive systems or the work conducted presents algorithmic perspective. In order to achieve the desired outcomes, co-evolving different kinds of mechanisms with different levels of complexity in their controllers is required. The research direction is the development of a mechanism, which just focus on the desired mechanism that can be co-evolved along with their controllers.

*3.4 Evolutionary robotics for physical prototyped*

ER for physical prototyped is not new, but very few researchers attempted it to develop the prototype using 3D printing method and compared the results of the evolved simulated robots to the prototyped one. Table IV shows the works conducted so far with real creature's development using ER. Pollack et al. [64] utilized evolutionary process to evolve the physics of robots, which is then prototyped through 3D printing machine to examine the performance [48]. This research work showed the applicability of an evolutionary process from the design to manufacturing. Lipson and Pollack [65] proposed elementary building blocks for both design and embodiment that showed a pathway to transfer of virtual diversity into reality. Authors [66] proposed a novel concept of the evolution of things, which was then utilized in various applications such as robot companion, environment friendly organism and evolutionary 3D printing. EvoFab – a 3D printing machine was developed to fabricate the robots, which has evolved through an evolutionary process [67] [68]. Though the research conducted in physical prototype are excellent as reported in Table IV. Further investigations are needed to analyze the proposed concepts for much realistic work.

## 4. Summary and analysis

This section presents the summary of various scientific literatures were presented on ER. The literatures are classified and organized in tabulated form for a quick review. Table I shows the applicability of evolutionary approaches in robot controller design. Literature related to robot design is presented in Table II. Table III and Table IV, respectively present the applicability of evolutionary approaches for robot morphology and control and for physical prototyping of robots. This summary and classification of existing research will be useful for the researchers to understand the importance of evolutionary approaches in the robotics.

**Table I.** Evolutionary robotics in evolutionary robot controller design.

| Citation | Author (s) | Year | Method used with main role |
|---|---|---|---|
| [6] | Beer and Gallagher | 1192 | GA with a continuous time RNN was utilized to evolve an effective neural controller for six legged insects like robot. |
| [7] | Ram et al. | 1994 | Navigation problem was divided into basic behaviors and then GAs was utilized to learn local robot navigation behavior for reactive control (collision free path). |
| [8] | Floreano and Mondada | 1994 | An evolutionary development (GA) of a real, NN driven mobile (Khepera) robot was presented. |

| [9] | Grefenstette and Schultz | 1995 | A SAMUEL-GA system was proposed that facilitate a collision free and the navigation task for mobile robots (Nomad 200). |
|---|---|---|---|
| [11] | Nolfi and Parisi | 1995 | A control system was proposed to perform a non-trivial sequence of behavior through an evolutionary process for mobile robots. |
| [10] | Meeden | 1996 | An incremental approach was proposed to evolve a NN controller for a 4-wheeled robot using GA. |
| [28] | Baluja | 1996 | A population based incremental learning was proposed for NN control in a robot. |
| [26] | Nordin and Banzhaf | 1997 | A GA was utilized for obstacle avoidance and object following behavior for Khepera. |
| [27] | Smith | 1997 | EA was implemented to evolve a vision based football playing Khepera. |
| [29] | Jeong and Lee | 1997 | A GA was proposed to evolve a two stage controller for two wheeled soccer playing robots. |
| [30] | Pratihar et al. | 1999 | A GA-fuzzy approached was proposed to evolve a fuzzy logic controller for collision free navigation in the presence of moving obstacles. |
| [59] | Watson et al. | 1999 | Proposed Embodied Evolution (EE) for automatic design of robotic controller. EE avoids the pitfalls of the simulate-and –transfer method, speedup of evaluation time through parallelism and well suited for multi-agent behaviors. |
| [31] | Pratihar | 2003 | EAs were proposed for a NN controller for the robot. |
| [32] | Karlra and Prakash | 2003 | A Neuro-GA based method was suggested for the inverse kinematics solution of a planner robotic manipulator. |
| [33] | Pires et al. | 2004 | MOGA was utilized to generate robot trajectory planning for 2R and 3R manipulator. |
| [34] | Nelson et al. | 2004 | The integrated ER environment was suggested for maze exploration. |
| [35] | Harvey et al. | 2006 | ER was suggested as a new scientific tool to study cognition. |
| [36] | Baldassarre and Nolfi | 2009 | Direct and evolutionary approaches were presented with their strengths and weaknesses in the development of the controller of autonomous robots. |
| [37] [38] | Koos et al. Koos et al. | 2010, 2013 | Fitness function and transferability was evolved to resolve the reality gap. The approach was tested for 8-dof quadrupedal robot. |
| [39] | Fukunaga et al. | 2012 | The GP based approach was implemented to evolve controller that performs high level tasks on service robots. |
| [40] | Risi and Stanley | 2013 | It evolves a special function using Hyper-NEAT. This approach takes morphology as input and produces NN controller fitted as output for the specific morphology. |

| Citation | Author(s) | Year | Utility of Evolutionary Robotics |
|---|---|---|---|
| [41] | Morse et al. | 2013 | Single-unit pattern generators (SUPG), which is indirectly encoded by a compositional pattern producing network (CPPN) evolved by HyperNEAT was used to evolve legged motion. |
| [42] | Chen and Hsieh | 2013 | It proposes a multilayer feed forward neural network based on the constructive approach using a GA. |

**Table II.** Evolutionary robotics in design.

| Citation | Author (s) | Year | Utility of Evolutionary Robotics |
|---|---|---|---|
| [43] | Chedmail and Ramstein | 1996 | It uses a GA for an open lopp robot for morphology and follows a given trajectory among obstacles. |
| [57] | Boudreau and Turkkan | 1996 | Proposed a floating point GA to solve the forward kinematic problem for parallel manipulators (two 3-degree-of-freedom planar and a 3-degree-of-freedom spherical) |
| [44] | Chocron and Bidaud | 1997 | Two level GA was proposed for the task based design of 3D modular manipulator. |
| [45] | Chung et al. | 1997 | A task based design method of modular robot manipulators was proposed which works in two steps: kinematic equations are used to determine the necessary configuration of the robot and then determine the optimal link length using the proposed efficient genetic algorithm. |
| [60] | Cheng and Lin. | 1997 | A GA is applied to obtain the optimal design of the biped controller and gait. |
| [46] [47] | Funes and Pollack, Funes and Pollack | 1998 1999 | A GA was used to evolve different 2D/3D morphologies using LEGO. |
| [48] [49] [50] | Hornby and Pollack Hornby et al. Hornby et al. | 2001, 2001, 2003 | The generative encoding mechanism was utilized to evolve physical, modular, 2D locomotive robot. |
| [56] | Asada et al. | 2001 | Cognitive developmental robotics (CDR) were proposed to design of humanoid robots. |
| [51] | Zykov et al. | 2007 | Autonomous self-reproduction capability of the robot was demonstrated. |
| [61] | Mucientes et al. | 2007 | Proposed a automated design of a fuzzy controller utilizing GAs to implement the wall following behavior in a mobile robot. |
| [58] | Jamwal et al. | 2009 | Proposed a soft parallel robot (SPR) for ankle joint rehabilitation. A global conditioning number (GCN) was defined through the Jacobian matrix as a performance index to determine the robot design. Finally, a modified GA was proposed to minimize GCN. |

| [52] | Rout and Mittal | 2010 | A differential evolution algorithm was suggested for 2-dof RR planar and 4-dof SCARA manipulator. |
| --- | --- | --- | --- |
| [53] | Bongard | 2011 | Robot morphology was evolved in the form of legs and an algorithm gait was proposed for legs. |
| [54] | Rubrecht et al. | 2011 | GA was implemented to evolve the design of a manipulator working in a highly constrained workspace for the maintenance of TBM in hostile conditions. |
| [55] | Tolley et al. | 2011 | An automated approach for stochastic modular robotics was presented that automatically evolve the target structure based on functional requirements. It showed the ability to evolve a structure and assembly plan to achieve a target function. |
| [42] | Chen and Hsieh | 2013 | A systematic approach was suggested for the rapid hardware synthesis of embedded real-time complex system. |

Table III. Evolutionary robotics for robot morphology and control.

| Citation | Author (s) and Year | Year | Utility of Evolutionary Robotics |
| --- | --- | --- | --- |
| [5] [62] | Sims, Sims | 1994 | Proposed a novel system to create virtual creatures in 3-dimensional physical world. Genetic language was presented that uses nodes and connections represent directed graphs, which are utilized to describe the morphology and neural circuitry. |
| [70] | Lee et al. | 1996 | Proposed a hybrid GP/GA approach to evolve combine morphology and control to achieve behavior-specified tasks such as walking, swimming and jumping. |
| [71] | Lund | 1997 | It includes the robot body plan in the genotype (as Hox gene) and then co-evolve a task fulfilling behaviors of body parameters in the morphological space using a GA. |
| [72] | Mautner and Belew | 2000 | Both hardware and the control structure were considered as variable in the evolution process using NNs. |
| [73][80] | Chocron and Bidaud , Chocron and Bidaud | 1999, 1999 | Proposed a method for the design of the complex mechatronic system using an evolutionary algorithm, which integrate a dynamic simulation of the robotic system in its environment. |
| [69] | Lipson and Pollack | 2000 | Proposed evolutionary approach to design of articulated bodies and fabricated the physical model through a 3D printing machine, which was then compared with the physical models. |
| [81] | Pollack and Lipson | 2000 | Undertaken a GOLEM project, which extends evolutionary approaches into the physical world. It evolves diverse electro- |

| | | | mechanical machines (robots). This research goes beyond the evolution of hardware controller. It demonstrated a path for the first time that allows transfer of virtual diversity of morphology into reality. |
|---|---|---|---|
| [82] | Paul and Bongard | 2001 | It utilizes GA and NN to achieve stable bipedal locomotion through the coupled evolution of morphology and control of a 5-linked biped robot in a physics based simulation environment. |
| [48] | Hornby and Pollack | 2001 | The generative encoding mechanism was utilized to evolve physical, modular, 2D locomotive robot. |
| [83] | Pollack et al. | 2001 | It presents a review of LegoBot, GOLEM and Thinkerbots. It also touched upon various robot design issues such as: automatically designed static structures, automatically designed and manufactured dynamic electromechanical systems and modular robots automatically using a generative DNA like coding. |
| [84] | Bongard and Lipson | 2004 | Proposed a new co-evolution approach referred as estimation-exploration algorithm, which automatically adapts the robot simulator utilizing the behavior of the target robot. This approach has shown several benefits: it automates the process of simulator and controller evolution, requires a minimum of hardware trials on the target robot, could be used in conjunction with other approaches to automate behavior transferable from simulator to reality. |
| [85] | Chocron et al. | 2005 | This research utilizes evaluation by dynamic simulation and optimization by artificial evolution and proposes a global design approach for self-reconfigurable locomotion system. Its primary objective of this research was to get a fully integrated robotic solutions in terms of morphology (topology and kinematics) and control (architecture and command). |
| [86] | Lipson | 2005 | Discusses the review of the research conducted in robots using the ER. |
| [87] [88] [89] [90] [91] [92] [93] [94] | Bongard, Nolfi and Floreano, Nolfi and Floreano, Meyer et al., Cliff et al., Harvey et al., Jakobi, Nelson. | 2013, 2000, 2000, 1998, 1993, 2005, 1997, 2009 | It presents a comprehensive review on evolutionary robotics. |

**Table IV.** Evolutionary robotics for physical prototyping of robot.

| Citation | Author (s) | Year | Utility of Evolutionary Robotics |
|---|---|---|---|
| [69] | Lipson and . Pollack | 2000 | Proposed evolutionary approach to design of articulated bodies and fabricated the physical model through a 3D printing machine, which was then compared with the physical models. |
| [67] | Rieffel and Sayles | 2010 | Proposed EvoFeb – a fully embodied evolutionary fabricator, was capable of producing novel objects in situ and it opens the door to a wide range of incredible exciting evolutionary design domain. |
| [68] | Kuehn | 2012 | Proposed evolutionary fabrication, which evolves a process rather than a product. It automatically invents and build anything from soft robots to new toys. |
| [66] | Eiben | 2012 | Proposed a new concept referred as an Evolution of Things, which leads to a new field of Embodied Artificial Evolution (EAE). |

## 5. Future research direction

Based on this brief survey, certain key areas of future investigation can be identified. In this section, author highlight some of the future research areas where further investigation is needed:

- The aforementioned discussion revealed that the research works conducted through the ERs are based on both evolution through generations and learning of individual entities. In addition, we noted that a high level of learning is achieved with ER techniques. Hence, an interaction between learning method and evolution with their effects on each other is likely to be studied to enhance the adaptive power of evolutionary processes.
- A detailed investigation is indicated in the field of evolutionary control design to improve the workability and efficacy analysis to design a flexible neural controller.
- Adaptive controller design methods play a major role in the modular robotic mechanism. But, without the use of adaptive controller developing a robotic mechanism to fulfill certain task requirements is highly challenging problem and a serious investigation is required to address this.
- So far, co-evolution or morphology has been utilized for robotic design. A further research would help in exploring the possibilities for co-evolution of desired robotic structure and its assembly planning.
- For a human being emotions play a significant role during decision making along with intelligence. Existing scientific literatures have shown connections between the aspects of artificial emotions and intelligence. A rigorous research and a mucg better rational thinking is required to apply evolutionary processes to understand and implement different aspects of a robotic system to evolve its appropriate cognitive architecture.

## 6. Conclusions

In this paper, the importance of evolutionary approaches for the robotic application development have been presented. Author have revisited the contributions made by various researchers in the implantation of evolutionary approaches in the robot controller design, robot body design, co-evolution of the body and brain and in transforming an evolved robot in physical reality. The summary of reviewed literatures have been shown in tabulated form for a quick review. Although, author have reported several research contributions that had already been done in the area of the ER. But, still there is scope for the further development of an advance and efficient approach for the betterment of the robot controller design, robot body design and others. This study have revealed that robotic design, development and control of robots are always remained  challenging. Hence, these research areas require further investigation for the improvement. Author have highlighted various challenges which creates room for further research in future.